\documentclass[10pt,twocolumn,letterpaper]{article}
\usepackage{cvpr}
\usepackage{times}
\usepackage{epsfig}
\usepackage{graphicx}
\usepackage{amsmath}
\usepackage{amssymb}
\usepackage{algorithm}  
\usepackage{algpseudocode}  
\usepackage{multirow}
\usepackage{booktabs}
\usepackage{bbding}

\usepackage[pagebackref=true,breaklinks=true,colorlinks,bookmarks=false]{hyperref}

\cvprfinalcopy 

\def\cvprPaperID{2604} 

\ifcvprfinal\fi
\begin{document}

\title{Upsampling Autoencoder for Self-Supervised Point Cloud Learning}

\author{Cheng Zhang, Jian Shi, Xuan Deng, Zizhao Wu$\thanks{Corresponding author.}$\\
Hangzhou Dianzi University, Hangzhou China\\
{\tt\small \{zhangcheng828,wuzizhao\}@foxmail.com,@hdu.edu.cn}}
\maketitle
\begin{abstract} 
In computer-aided design (CAD) community, the point cloud data is pervasively applied in reverse engineering, where the point cloud analysis plays an important role. 
While a large number of supervised learning methods have been proposed to handle the unordered point clouds and demonstrated their remarkable success, their performance and applicability are limited to the costly data annotation.   
In this work, we propose a novel self-supervised pre-training model for point cloud learning without human annotations, which relies solely on upsampling operation to perform feature learning of point cloud in an effective manner. 
The key premise of our approach is that upsampling operation encourages the network to capture both high-level semantic information and low-level geometric information of the point cloud, thus the downstream tasks such as classification and segmentation will benefit from the pre-trained model. 
Specifically, our method first conducts the random subsampling from the input point cloud at a low proportion e.g., 12.5\%. Then, we feed them into an encoder-decoder architecture, where an encoder is devised to operate only on the subsampled points, along with a upsampling decoder is adopted to reconstruct the original point cloud based on the learned features. Finally, we design a novel joint loss function which enforces the upsampled points to be similar with the original point cloud and uniformly distributed on the underlying shape surface.
By adopting the pre-trained encoder weights as initialisation of models for downstream tasks, we find that our UAE outperforms previous state-of-the-art methods in shape classification, part segmentation and point cloud upsampling tasks. Code will be made publicly available upon acceptance.  
\end{abstract}


\section{Introduction}
\label{sec:intro}
Analysis and classification of 3D point cloud is an important problem in computer vision, graphics and CAD communities, due to its wide applications in robot manipulation \cite{RusuMBDB08}, autonomous driving \cite{QiLWSG18}, reverse engineering \cite{BeniereSGBP13}, etc. 
In view of the great success of deep learning on other computer vision tasks, many endeavors have been made to adapt the deep learning technologies to the analysis of 3D point clouds \cite{2018Dynamic,qi2017pointnet,qi2017pointnet++,2018PointCNN,2020Point,2020PointASNL,xu2020grid,guo2020pct,zhao2020point,patchformer,pvt} including PointNet \cite{qi2017pointnet}, VoxNet \cite{zhou2018voxelnet}, etc. However, most of the existing works are supervised and rely on large-scale and accurate annotated dataset, which hinder their applicability. 

Unsupervised pre-training, on the other hand, has emerged as a promising alternative to address the shortcoming of the above supervised methods within an effective architecture. Depending on the models their used, existing unsupervised approaches for learning on point clouds can be roughly classified into two categories: reconstruction-based and generation-based methods. The generation-based methods \cite{MAP-VAE,LGAN} typically employed the Generative Adversarial Networks (GANs) \cite{LGAN} or Variational Auto-Encoders (VAEs)\cite{MAP-VAE} to learn feature representations in an unsupervised framework. 
The reconstruction-based models \cite{sauder2019self} usually adopted a framework that trains an encoder to learn shape representations by reconstructing the input data via a decoder.
Though these models have been demonstrated to be effective in certain applications, they usually fail to acquire high-level structure information \cite{chen2021unsupervised}. 


\begin{figure}
    \centering
    \includegraphics[width=\linewidth]{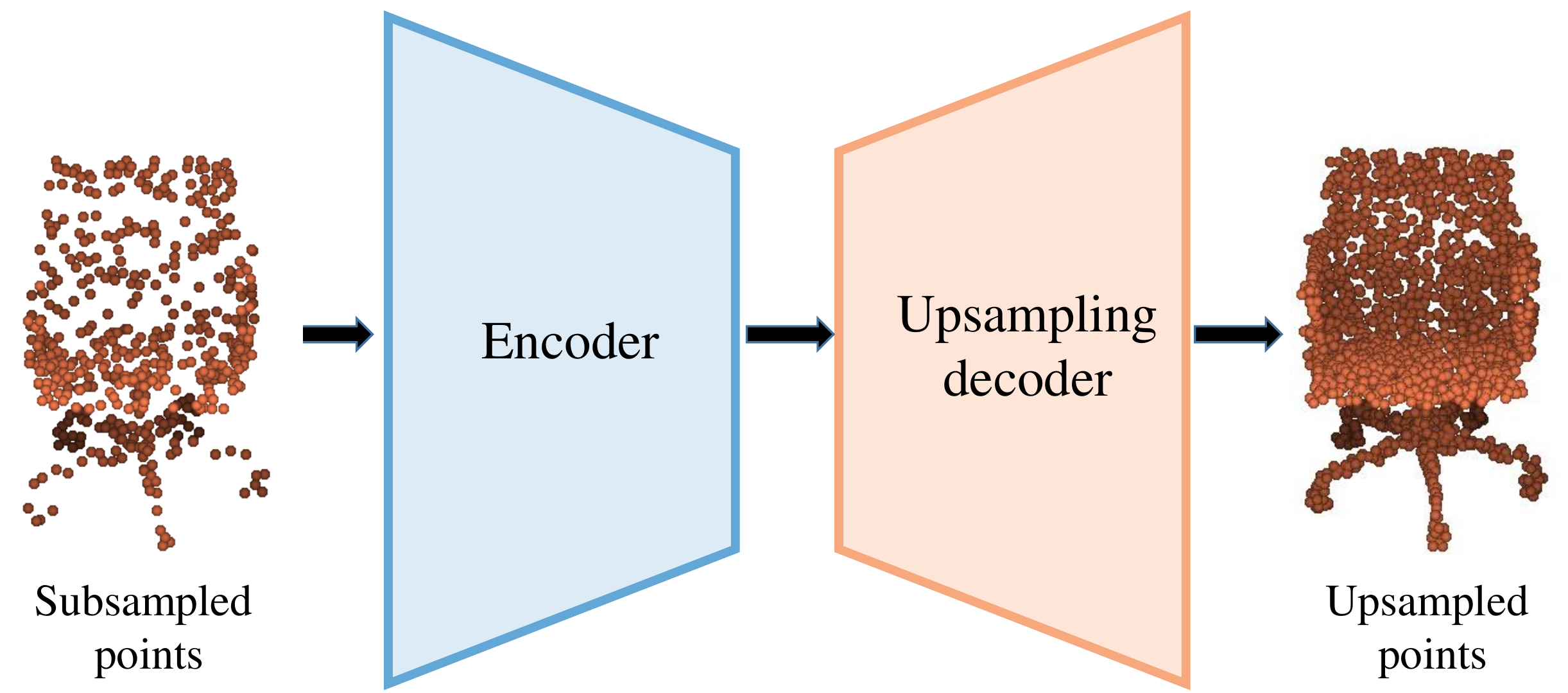}
    \caption{Overview of UAE. During pre-training, we first generate subsampled points for each input by randomly sampling, then train a encoder-decoder model to upsample these points by 8$\times$. Finally, we use the learned pre-training weights as an initialization for various downstream tasks.}
    \label{overview}
\end{figure}

Recently, self-supervised pre-training \cite{sun2021point,gu2021staying,sauder2019self,rao2020global,zhang2021self,chen2021unsupervised}, a class of unsupervised pre-training methods, is gaining certain interest in the point cloud learning paradigm. The premise of self-supervised learning is to generate the results from unlabeled data in a supervised manner by leveraging the underlying structure in the data. By adopting the pre-trained self-supervised models for point cloud learning, these approaches are capable of predicting high-level structure information and with good performance. 
For example, some self-supervised pre-training methods\cite{xie2020pointcontrast,jiang2021unsupervised,du2021self} have employed the contrastive learning framework to capture high-level geometric information (e.g., shape representation) by modeling shape similarity and dissimilarity between two or more views. OcCo \cite{wang2021unsupervised} has utilized occlusion completion as a pre-training task to learn pre-train weights for point cloud analysis. However, according to our knowledge, all the existing self-supervised pre-training approaches are not efficient enough, as the majority of them need a careful treatment of negative pairs by either relying on large batch sizes \cite{xie2020pointcontrast,zhang2021self}, memory banks \cite{jiang2021guided}, data pre-processing \cite{wang2021unsupervised} or customized mining strategies \cite{du2021self,jiang2021guided} to retrieve the negative pairs. Furthermore, their performance critically depends on complicated 3D data augmentations, e.g., cuboid \cite{zhang2021self}, shape disorganizing \cite{chen2021shape} and shearing \cite{gao2020graphter}.


In this work, we propose a novel self-supervised pre-training model based on upsampling auto-encoder, namely UAE, aiming to reach a simple yet effective model, which can be applied to a wide range of downstream point cloud analysis tasks such as 3D object classification, semantic segmentation and point cloud upsampling. The key observation is that upsampling model that is accurate needs to understand the structure information of the point cloud, and thus a pre-trained upsampling model can facilitate some downstream tasks of point clouds due to the high-level structure information it inherently owned. 
We note that some works have explored the upsampling or completion strategies for self-supervised point cloud learning, however as mentioned former, these approaches require either negative sampling or data augmentation operation. Our UAE is inspired by the masked auto-encoder (MAE) \cite{he2021masked} for image analysis, offering a simpler and more effective upsampling architecture than theirs by providing a solution that subsamples random points from the input point cloud and reconstructs the points that are not sampled in the point space directly. 
Specifically, our method first conducts the random subsampling from the input point cloud at a low proportion. Then, we feed them into an asymmetric encoder-decoder architecture, where the encoder is devised to operate only on the subsampled points, along with an upsampling decoder is adopted to reconstruct the original point cloud based on the learned features. Finally, we design a novel joint loss function that enforces the upsampled points to be similar with the original point cloud and uniformly distributed on the underlying shape surface.

We pre-train our UAE network with ShapeNet dataset \cite{chang2015shapenet} and evaluate its performance in various downstream point cloud learning tasks, including shape classification, shape segmentation and point cloud upsampling. Our experiments have demonstrated that in all these tasks, our pre-trained model offers better performance than the same network trained with the labeled data of downstream tasks. And comparing with other SOTA unsupervised pre-training methods, our UAE obtains the best performance in nearly all these tasks. 

Our contributions can be summarized as follows:
\begin{itemize}
    \item presents a novel self-supervised pre-training method for point cloud learning based on pre-trained upsampling model. 
    \item investigates a novel upsampling architecture, where a joint loss function is introduced to enforce the upsampled points to be similar to the original point cloud.
     \item achieves significant performance in shape classification, part segmentation and point cloud upsampling tasks.
\end{itemize}


\section{Related works}
\label{sec:related}
\subsection{Supervised Learning on 3D Point Clouds}
In recent years, deep learning on 3D point cloud has attracted increasing attention of researchers. As 3D points are irregularly sampled and have quite a few special properties such as irregularity and permutation invariance,  the traditional neural networks, i.e., convolutionl neural networks (CNNs) in the 2D field, cannot be directly applied to point clouds.
Therefore, previous works attempt to convert point clouds into a regular grid structure and apply 3DCNN to feature learning \cite{zhou2018voxelnet,xu2020grid,2019VoxSegNet}. However, the computational consumption and memory of 3DCNN increase cubically with the increase of voxel’s resolution \cite{2019Point}.

Another mainstream method is directly operating the unordered point clouds \cite{2018Point,2018Point2Sequence,2019PointWeb,guo2020pct,zhao2020point,2019Graph,graphpbn,2020ClusterNet,2021Semantic}. For instance, Qi et al. \cite{qi2017pointnet} pioneered this series of works and proposed a novel neutral architecture, PointNet, which can directly handle irregular and unordered 3D points by stacking multi-layer perceptrons (MLPs) and capturing global shape with max-pooling. Later, Qi et al. \cite{qi2017pointnet++} proposed PointNet++ to overcome the disadvantage that PointNet fails to capture local structures by developing a hierarchical grouping architecture at different set abstraction level. Subsequently, PointCNN \cite{2018PointCNN}, PointConv \cite{wu2019pointconv}, RSCNN \cite{RSCNN} and DGCNN \cite{2018Dynamic} also focus on local structures of point cloud and further improve the quality of captured features. However, these models mentioned above are supervised and rely on large-scale labeled point cloud dataset. In this paper, we propose a novel unsupervised framework, namely UAE for point cloud analysis.

\subsection{Unsupervised Learning on 3D Point Clouds}
There are several prior attempts \cite{MAP-VAE,achituve2021self,achlioptas2018learning,gadelha2018multiresolution,hassani2019unsupervised,foldingnet} on learning shape-specific invariant representations of a point cloud based on unsupervised reconstruction models. These methods discover valuable information of 3D point cloud by reconstructing the input data, which has been shown to effectively learn shape-specific invariant representations. For example, FoldingNet \cite{foldingnet} proposed an end-to-end auto-encoder to get the codeword that can represent the high-dimensional embedding of a point cloud, and the fully-connected decoder was replaced with the folding-based decoder. GraphTER \cite{gao2020graphter} proposed a novel unsupervised learning method to capture intrinsic patterns of point cloud structure under both global and local transformations.
However, because of lacking sufficient semantic supervision, this kind of methods are insufficient to understand the high-level semantic information of point clouds. 

Witnessing the great success of self-supervised on computer vision tasks, many researchers have endeavored to apply it to the field of self-supervised point cloud analysis. Xie et al. \cite{xie2020pointcontrast} pioneered this line of works by proposing PointContrast to build representations of scene-level point cloud which relies on the complete 3D reconstruction of a scene with point-wise correspondences between the different views of a point cloud. However, these point-wise correspondences need to post-process the input data by registering the different depth maps into a single 3D scene. Later, Zhang et al.\cite{zhang2021self} proposed DepthContrast to side-step the necessity of registered point clouds or correspondences, by considering each depth map as an instance and discriminating between them, even if they come from the same scene. Yet, most of them require careful treatment of negative key samples by either relying on large batch sizes \cite{xie2020pointcontrast,zhang2021self}, memory banks or customized mining strategies \cite{du2021self} to find the negative key samples. Additionally, the performance of these methods critically depends on the choice of data augmentations \cite{xie2020pointcontrast,gu2021staying,gao2020graphter,jiang2021unsupervised}, limiting the applicability of the approach. 
As a comparison, in this work, we investigate a useful pretext task for learning shape-specific invariant representations. 

Some researchers have also exploited the self-supervised pre-training model, for example,  OcCo \cite{wang2021unsupervised} is investigated by using occlusion completion as a pre-training task to learn an initialization for point cloud models.
Compared with OcCo, our model has two advantages: firstly, our UAE does not require extra data-preprocessing to generate occluding point clouds based on different viewpoints; and secondly, our method uses point cloud upsampling as a pre-training task and design a novel joint loss function to capture better shape information by enforcing the upsampled points uniformly distributed on the underlying shape surface.


\begin{figure*}[t]
    \centering
    \includegraphics[width=\linewidth]{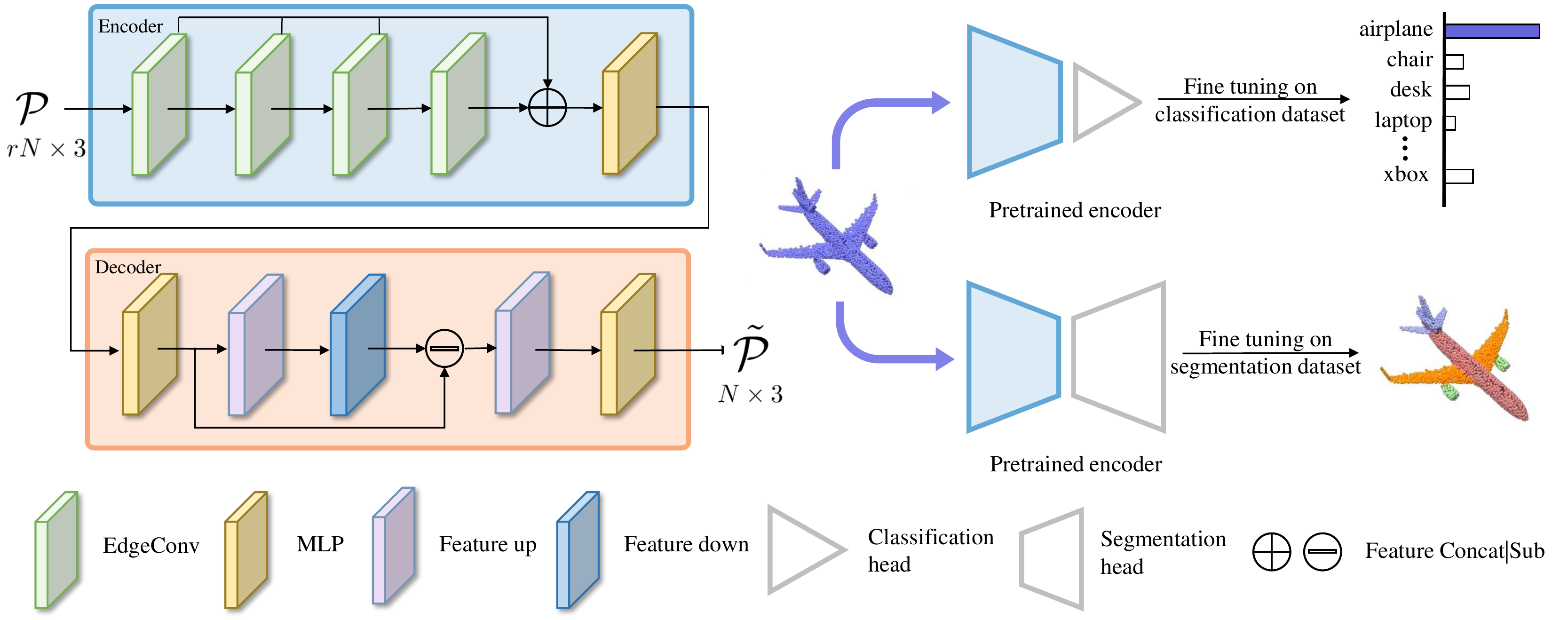}
    \caption{\textbf{Model Structure.} The UAE is mainly composed of one encoder and one decoder. \textbf{Encoder:} The encoder takes $rN$ points as input, and aggregates the local features of each point at an EdgeConv layer. To obtain multiple-scale features, we concatenate the feature maps from shallower EdgeConv layers and feed them into a MLP layer as the output of the encoder. \textbf{Decoder:} The decoder consists of a MLP layer, a feature up block, and a feature down block which is shown in Figure \ref{fig:twoblocks}. The decoder upsample $\mathcal{P}$ to the original resolution $\tilde{\mathcal{P}}$, and minimize the chamfer distance between $\mathcal{X}$ and $\tilde{\mathcal{P}}$. \textbf{Downstream tasks. }we use the learned encoder weights as initialization for various downstream tasks, e.g., shape classification, part segmentation and point cloud upsampling.}
    \label{fig:framework}
\end{figure*}

\subsection{Self-Supervised Learning for Images}

Self-supervised learning has attracted increasing attention since the cost of accurate annotated dataset is extraordinarily expensive \cite{mahajan2018exploring,olshausen1996emergence,salakhutdinov2009deep,vincent2008extracting}. There are several classes of methods for learning representations such as clustering \cite{bojanowski2017unsupervised,caron2018deep}, GANs \cite{donahue2016adversarial,mescheder2017adversarial}, pretext tasks \cite{doersch2015unsupervised,noroozi2016unsupervised} etc. Recently, contrastive learning \cite{oord2018representation,dosovitskiy2015discriminative,wu2018unsupervised,tian2020contrastive,he2020momentum,chen2020simple} have been proposed to learn the unsupervised representation of 2-dimensional natural images. It can be considered as a pretext task where the goal is to maximize the representational similarity between positive key samples and dissimilarity between negative key samples for the input query. The positive key sample is generated with a random data augmentation module, which, given the input, generates a pair of random views of the input. Other inputs in the same batch are often used as negative key samples. Generally, contrastive and related methods strongly depend on
data augmentation and the sampling of negative pairs \cite{he2021masked,byol}, which limit their applicability. In contrast to these methods, we pre-train the UAE on a simple but meaningful upsampling objective, aiming to capture high-level semantic information within a simple and effective framework. 

\section{Method}
Suppose that the original point cloud with $N$ points is denoted by $\mathcal{X}\in \mathbb{R}^{N\times C} $. In the simplest setting of $C$ = 3, each point contains 3D coordinates. We aim to train a model that is capable of unsupervised learning the point cloud representations. Towards this goal, we propose upsampling autoencoder (UAE), a simple self-supervised learning approach that reconstructs the original point cloud from a small number of subsampled points. Unlike classical self-supervised methods, our UAE is suggested to capture high-level semantic information without any data augmentation and negative pairs. The overall structure of UAE is shown in Figure \ref{fig:framework}.

In what follows, we begin by describing the strategy that makes up our subsampled points. Then we detail how to learn a encoder $\varphi(\cdot)$ and an upsampling decoder $\phi(\cdot)$. And finally, we discuss the design of our joint loss function.

\subsection{Subsampling}
Given a point cloud $\mathcal{X}\in \mathbb{R}^{N\times 3}$, we first subsample $\mathcal{X}$ into a lower resolution subset $\mathcal{P}\in \mathbb{R}^{rN\times 3}$, where $r$ denotes that subsampling ratio. There exists a variety of sampling methods: 1) Random Sampling, where the probability of sampling each point in the point cloud follows a uniform distribution; 2) Farthest Point Sampling, where each point to be sampled is as far away as
possible from points in the sampled set; 3) Local Sampling, where the points are sampled from a local part of the point cloud. In this paper, we propose to randomly sample a subset of points at a low proportion uniformly (e.g., $r=0.125$), which largely eliminates redundancy, thus creating a rather challenging task that cannot be easily handled by extrapolation from subsampled neighboring points \cite{he2021masked}. This highly sparse point cloud is more conducive for us to design an effective encoder to capture high-level semantic information.

\subsection{Encoder}
The encoder $\varphi(\cdot)$ takes the coordinates of the subsampled point clouds $\mathcal{P}$ as input and outputs high-dimensional features. In contrast to previous self-supervised methods, where the global shape representation is learned by the encoder, we perform point-wise feature extraction on subsampled point cloud ${\mathcal{P}}$, where each point can learn its representation by upsampling decoding the original point cloud to reveal the local structure around it. These representations will gather global shape information about the original point cloud as we randomly subsample points into different subsets (or groups) through training iterations while capturing the local structures under point-wise upsampling decoding.

Generally, any deep learning-based network that takes point clouds as the input and outputs high-dimensional features can be adopted as the encoder $\varphi(\cdot)$ of UAE. However, as 3D points have some special properties such as irregularity and permutation invariance, we can not directly leverage the convolutional neural network (CNN). Therefore, we deploy the dynamic graph CNN (DGCNN \cite{2018Dynamic}) to overcome this limitation. In particular, we adopt the EdgeConv layer in DGCNN as our basic feature extraction block. By performing EdgeConv, our encoder can aggregate the features of neighbor points to the center point and update the feature of the center point. 

Specifically, given an input subsampled points $\mathcal{P}= \{p_i\}_{i=1}^{rN} \subseteq \mathbb{R}^C$, we first apply K-Nearest Neighbor (KNN) algorithm to construct a local graph in the feature space. The KNN algorithm based on the feature space can efficiently and effectively find two points with the most similar semantics, i.e., the points on the two wings of the airplane, which have similar semantics. Then, the EdgeConv layer performs feature transformation on the $rN$ points and output the updated point features of $p_i$. The output feature of a point $p_i$ is
\begin{equation}
    f_i = \max_{p_j\in \mathcal{N}(p_i)}{\rm ReLU}(h_{\theta} (p_i,p_i-p_j)),
\label{edgeconv}
\end{equation}
where $p_j\in \mathcal{N}(p_i)$ denotes that point $p_j$ belongs to the neighborhood of point $p_i$, $h_{\theta}$ denotes that the learnable parameters of MLP.

By stacking four EdgeConv layers, each point is aggregated with local neighboring features hops away. Furthermore, as the subsampling operation enlarges the distance between neighboring points, performing EdgeConv for the subsampled points will capture longer-range dependencies and thus facilitates to extract high-level semantic information that is invariant under random sampling.

\subsection{Decoder}
\begin{figure}
    \centering
    \includegraphics[width=\linewidth]{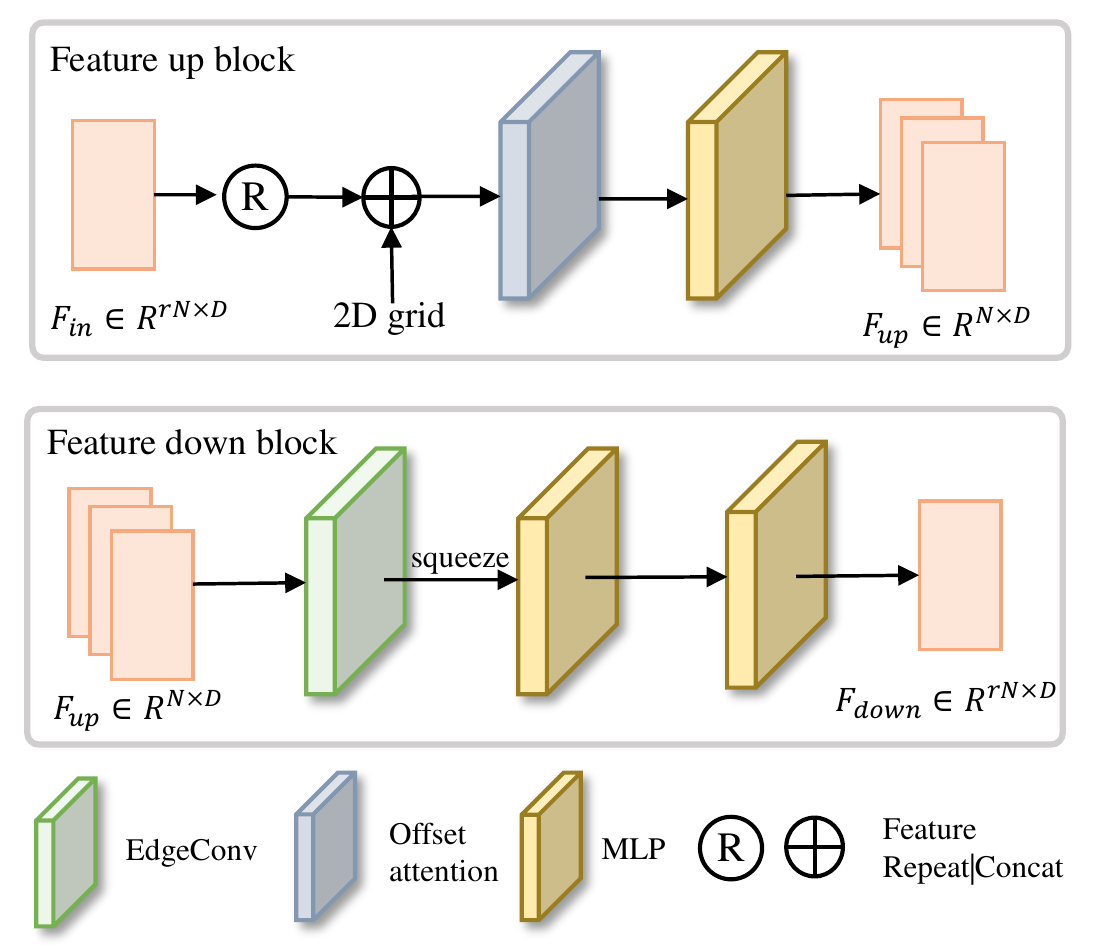}
    \caption{\textbf{The overview of feature up (top) and feature down (bottom) blocks}, where $r$ is the subsampling ratio and $D$ is the dimension of point feature. We construct our upsampling decoder based on these blocks.}
    \label{fig:twoblocks}
\end{figure}

The decoder $\phi(\cdot)$ takes the point-wise features as input and outputs the upsampled point cloud $\tilde{\mathcal{P}}$.
To upsample a point cloud, Li et al. proposed PU-Net \cite{PUNet} to duplicate the point features and then employ separate MLPs to process each copy independently. However, the expanded features would be too similar to the inputs, so affecting the upsampling quality. In recent years, wang et al. proposed MPU \cite{MPU} to break a $16\times$ upsampling network into four successive $2\times$ subnets to progressively upsample points in multiple steps. MPU preserves superior upsampling details but usually requires a more complex training process.

Inspired by the PU-GAN \cite{PUGAN}, we design a novel upsampling decoder to expand the point features, which is mainly composed of feature up and feature down blocks. As shown in Figure \ref{fig:framework}, we first upsample the point features $F_{in}$ (after a MLP layer) by $\frac{1}{r}\times$ to generate $F_{up}$ and downsample it to generate $F_{down}$; then, instead of directly constructing the original point cloud $\mathcal{X}$, we adopt residual learning to regress the per-point feature offset by calculating the difference between $F_{up}$ and $F_{down}$; Ultimately, we feed them into a feature up block and a MLP layer to restore the original point cloud $\mathcal{X}$.
Such a strategy that utilizes feature offset to self-correct the expanded features has two advantages: firstly, it facilitates the production of fine-grained features while avoiding tedious multi-step training; secondly, the features of the same object can be completely different with respect to rigid transformations. 
In the following, we detail the design choices of feature up and feature down blocks. 

\textbf{Feature up block. }To upsample the point features $\frac{1}{r}$ times, we adopt the commonly-used variation expansion operator \cite{PUGAN,Li_2021_CVPR} by duplicating $F_{in}$ with $\frac{1}{r}$ copies and concatenating with a regular 2D grid. However, such operator may introduce redundant information or extra noise \cite{Li_2021_CVPR}. To rectify these problems, we propose to make use of the offset-attention \cite{guo2020pct} as the global refinement unit by considering
the overall shape structure. The reason behind is that, compared with the widely-used self-attention unit \cite{PUGAN,Li_2021_CVPR,zhang2019self}, the offset-attention is generally more robust because it works by replacing the attention feature with the offset between the input of self-attention module and attention feature. 
The pipeline of this block is illustrated in Figure \ref{fig:twoblocks} (top).

\textbf{Feature down block. }To downsample the expanded features, we propose a novel GCN-based feature down block, illustrated in Figure \ref{fig:twoblocks} (bottom). Given the expanded features $F_{up}\in \mathbb{R}^{N\times D}$, our method works in two steps. First, we reshape $F_{up}$ of shape $N \times D$ to $rN \times\frac{1}{r}\times D$ and use one layer of EdgeConv to downsample $F_{up}$ to shape $rN \times D$ using learnable parameters. Second, we feed them into a set of MLPs to regress the point features $F_{down}$.

In contrast to previous works \cite{PUGAN,PUNet,MPU}, our feature down block leverages GCNs, which are common modules for feature extraction. To the best of our knowledge, we are the first to introduce a GCN-based feature downsampling block. Our GCN design choice stems from the fact that GCNs enable our feature down block to encode spatial information from point
neighborhoods and learn new features from the latent space rather than simply using Convs.

\begin{figure}
    \centering
    \includegraphics[width=\linewidth]{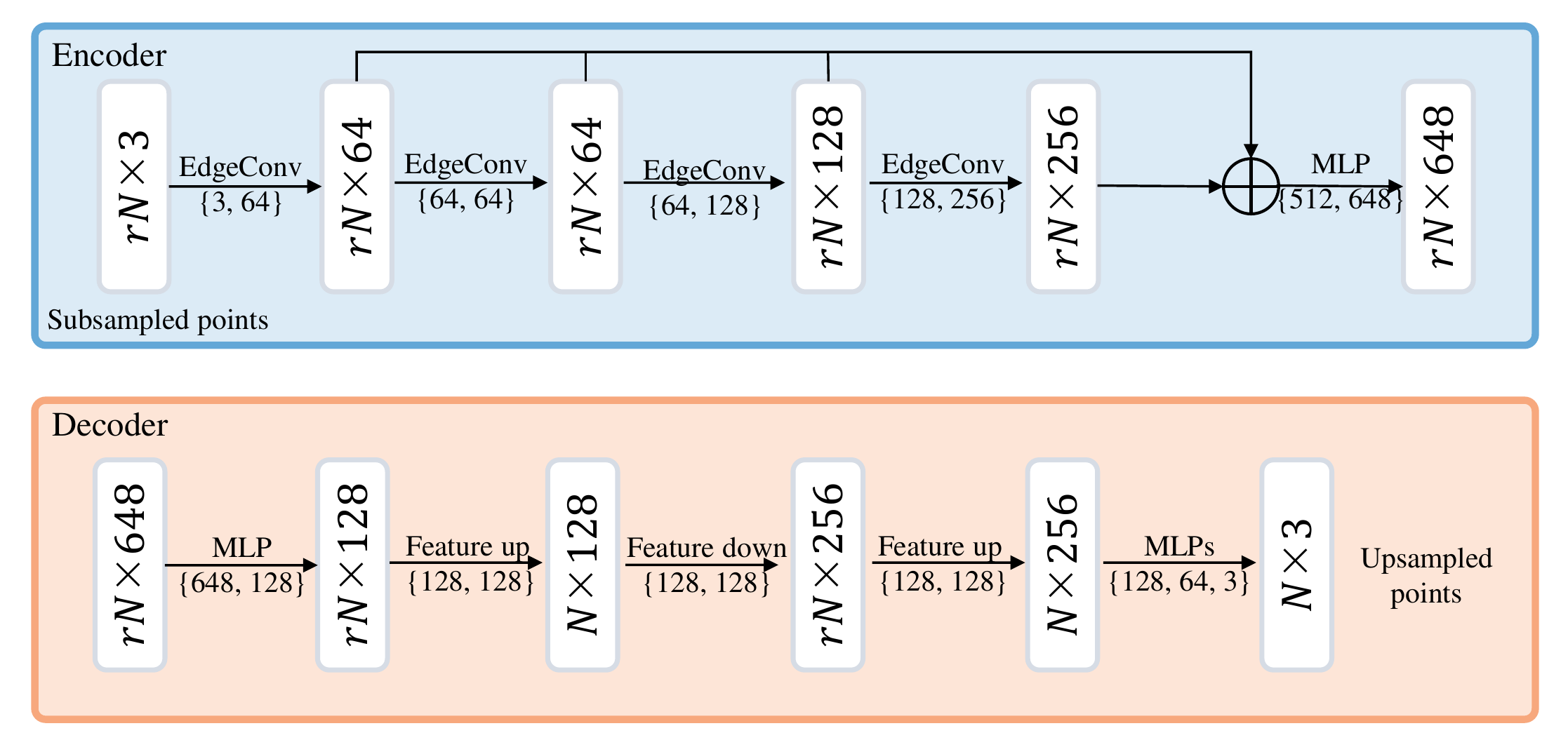}
    \caption{Pre-training network architecture of UAE. We adopt DGCNN \cite{2018Dynamic} architecture as the encoder $\varphi(\cdot)$ to map a subsampled point cloud into a 648-dimensional point-wise features. We design the decoder $\phi(\cdot)$ to take the point-wise features as input and output the upsampled point cloud $\tilde{\mathcal{P}}$.}
    \label{network_arch}
\end{figure}

\subsection{Joint loss function}
To encourage the upsampled points to be similar to the original point cloud and uniformly distributed on the underlying shape surface, a novel joint loss function was proposed to train our UAE in an end-to-end fashion. Next, we will detail the design of this loss function.

\textbf{Reconstruction loss.} Our UAE reconstructs the original point cloud $\mathcal{X}$ by predicting coordinate for each unsampled point. Each element in the decoder's output is a vector of coordinate values representing a point's spatial location. We formulate our objective function to encourage the geometric consistency between $\tilde{\mathcal{P}}$ and $\mathcal{X}$ in the point space:
\begin{equation}
\mathcal{L}_{CD}(\tilde{\mathcal{P}},\mathcal{X})=\frac{1}{|\tilde{\mathcal{P}}|}\sum_{\tilde{p}\in\tilde{\mathcal{P}}}\min_{x\in\mathcal{X}}||\tilde{p}-x||_2 + \frac{1}{|\mathcal{X}|}\sum_{x\in\mathcal{X}}\min_{\tilde{p}\in\tilde{\mathcal{P}}}||x-\tilde{p}||_2,
\label{chamefer_dist}
\end{equation}
where $\mathcal{L}_{CD}(\cdot)$ means the Chamfer Distance (CD) to measure the average closest point distance between two point sets, $||\cdot||_2$ denotes the L2 distance between two points.

\textbf{Repulsion loss.} Since the reconstruction loss alone can not ensure that the upsampled points over the underlying object surfaces will be uniformly distributed, which is important for capturing high-level shape information. To solve this issue, we propose to use the repulsion loss \cite{PUNet} as the constraint term of uniformity distribution, which is represented as:
\begin{equation}
\mathcal{L}_{rep} = \sum^{N}_{i=0}\sum_{\tilde{p_j}\in \mathcal{N}(\tilde{p_i})}\delta(||\tilde{p_j}-\tilde{p_i}||)\omega(||\tilde{p_j}-\tilde{p_i}||),
\label{repulsion_loss}
\end{equation}
where $\mathcal{N}(\tilde{p_i})$ is the point set of the k-nearest neighbors of point $\tilde{p_i}$, $N$ is the number of upsampled points and $||\cdot||$ is the L2-norm. $\delta(m)=-m$ is the repulsion term, which is a decreasing function to penalize $\tilde{p_i}$ if $\tilde{p_i}$ is located too close to other points in $\mathcal{N}(\tilde{p_i})$. We further use the fast-decaying weight function $\omega(m)= e^{-m^2/h^2}$ ($h$ as the finite support radius \cite{07h}) to penalize $\tilde{p_i}$ only when it is too close to its neighboring points.

\textbf{Overall loss function. }The overall loss function can be calculated as the weighted sum of all the terms described above:
\begin{equation}
    \mathcal{L}_{total} = \alpha\mathcal{L}_{CD}+\beta\mathcal{L}_{rep},
\end{equation}
where $\alpha$ and $\beta$ are the weighting factors for the loss functions of reconstruction and repulsion, respectively. Following \cite{PUNet}, we set $\alpha = 100$ and $\beta = 1$.
\section{Experiments}
In this section, we detail the pre-training setups of UAE and evaluate it for different downstream tasks: classification, part segmentation, and
point cloud upsampling.

\subsection{Experimental setups}
\textbf{Pre-training Dataset.}  We train our UAE on ShapeNet dataset \cite{chang2015shapenet}, which consists of 57, 448 synthetic 3D CAD models organized into 55 categories with a further 203 subcategories, organized according to WordNet synsets. For pre-training we use the normalized version of ShapeNet, where all shapes are consistently aligned and normalized to fit inside a unit cube.

\textbf{Architecture.} The network architecture used for pre-training is shown in Fig \ref{network_arch}. we deploy four EdgeConv layers whose dimensions are [64,64,128,256] and one MLP (648) layer as the encoder. The number of nearest neighbors set to 20 for all EdgeConv layers. After the four EdgeConv layers, we concatenate the output of these layers to get a 64+64+128+256=512 dimensional point cloud features and feed them into one MLP layer with input channels of 512 and output channels of 648. Similar to previous works \cite{PUGAN,pugcn}, the upsampling decoder is used to obtain the reconstructed point cloud, where $D$ is 128.

\textbf{Pre-training Hyperparameters.} 
We uniformly sample 2, 048 points on ShapeNet dataset for our UAE pre-training.
We use the Adam optimizer with no weight decay (L2 regularisation). The learning rate is set to 1e-3 initially and is decayed by 0.7 every 10 epochs. We pre-train the models for 120 epochs. The batch size is 32, and the momentum of batch normalization is 0.9. Note that, all the experiments are trained on an NVIDIA RTX3090 GPU.

\textbf{Evaluation Metrics.} For the classification task on ModelNet10 and ModelNet40 datasets, we use the overall accuracy (OA) as the metric. On ShapeNet Part dataset, we evaluate our scheme with part classification accuracy and mean Intersection over-Union (mIoU). For each sample, IoU is computed for each part that belongs to that object category. The mean of
all part IoUs is regarded as the IoU for that sample. For point cloud upsampling task, we use the Chamfer distance (CD), Hausdorff distance (HD), and point-to-surface distance (P2F) w.r.t ground truth meshes as evaluation metrics. The smaller the metrics, the better the performance.

\begin{table}[tb]
\centering
\resizebox{\linewidth}{!}{ 
\begin{tabular}{llcc}
\hline
Model & Reference & ModelNet40 & ModelNet10 \\
\hline
\hline
\multicolumn{4}{c}{Supervised Learning} \\
PointNet \cite{qi2017pointnet} & CVPR2016 &89.2 & $-$ \\
PointNet++ \cite{qi2017pointnet++} & NIPS2017& 91.9 & $-$  \\
PointConv \cite{2018PointConv}  & CVPR2019 & 92.5 & $-$   \\   
RGCNN \cite{2018RGCNN}  & MM2018 & 90.5 & $-$   \\
PointCNN \cite{2018PointCNN} & NIPS2018 & 92.2 & $-$  \\  
SpiderCNN \cite{2018SpiderCNN}  & ECCV2018 & 92.4 & $-$   \\
PointWeb \cite{zhao2019pointweb}  & CVPR2019 & 92.3 & $-$  \\  
DGCNN \cite{2018Dynamic} & TOG2018& 92.9 & $-$ \\
\hline
\hline
\multicolumn{4}{c}{Unsupervised Transfer Learning} \\
FoldingNet \cite{foldingnet} & CVPR2018 &88.4 & 94.4\\
MAP-VAE \cite{MAP-VAE} & ICCV2019& 90.1 & 94.8 \\
MID-FC \cite{wang2020unsupervised} & TOG2020& 90.3 & $-$ \\
GSIR \cite{chen2021unsupervised}  & ICCV2021& 90.3 & $-$ \\
RS-DGCNN  \cite{sauder2019self}  & NIPS2019& 90.6 & 94.5 \\
GLR-RSCNN \cite{rao2020global}  & CVPR2020 & 91.3 & 94.2 \\
GraphTER \cite{gao2020graphter} & CVPR2020& 92.0 & $-$ \\
PointDis \cite{PointDist}  & $-$& 92.3 & 95.3 \\
SSC-RSCNN \cite{chen2021shape}  & ICCV2021& 92.4 & 95.0 \\
Ours-DGCNN  & $-$& \textbf{92.9}& \textbf{95.6} \\
\hline
\hline
\multicolumn{4}{c}{Supervised Fine-tuning} \\
DepthContrast \cite{zhang2021self}& CVPR2021 &91.3 & $-$\\
MID-FC  & TOG2020& 93.1 & $-$ \\
GLR-RSCNN  & CVPR2020& 92.2 & 94.8\\
ParAE-DGCNN  & ICCV2021& 92.9 & $-$\\
SSC-RSCNN  & ICCV2021& 93.0 & 95.5 \\
Ours-DGCNN  & $-$& \textbf{93.2} & \textbf{95.7} \\
\hline
\end{tabular}
} 
\caption{
\textbf{Shape Classification Results on ModelNet40 and ModelNet10.} The quantitative results of SOTA unsupervised and supervised fine-tuning methods. “Unsupervised Transfer Learning” denotes the parameters of the pre-trained encoder are fixed on downstream tasks, and “Supervised Fine-tuning” denotes the pre-trained encoders are fine-tuned on target tasks.
} 
\label{tab:cls}
\end{table}

\subsection{Shape Classification}

\begin{table*}[tb]  
\centering
\begin{center}  
\begin{tabular}{p{1.8cm}p{0.5cm}p{0.5cm}p{0.5cm}p{0.5cm}p{0.5cm}p{0.5cm}p{0.5cm}p{0.5cm}p{0.5cm}p{0.5cm}p{0.5cm}p{0.5cm}p{0.5cm}p{0.5cm}p{0.5cm}p{0.5cm}p{0.5cm}}  
\hline
\footnotesize Model & \footnotesize mIoU & \footnotesize Areo & \footnotesize Bag & \footnotesize Cap & \footnotesize Car & \footnotesize Chair & \footnotesize Ear  \footnotesize Phone & \footnotesize Guitar & \footnotesize Knife & \footnotesize Lamp &  \footnotesize Laptop &  \footnotesize Motor & \footnotesize Mug & \footnotesize Pistol & \footnotesize Rocket & \footnotesize Skate \newline \footnotesize Board & \footnotesize Table\\ 
\hline
\hline
\multicolumn{18}{c}{Supervised Learning} \\
\footnotesize KDNet\cite{klokov2017escape} & 82.3 & 80.1 & 74.6 & 74.3 & 70.3 & 88.6 & 73.5 & 90.2 & 87.2 & 81.0 & 94.9 & 57.4 & 86.7 & 78.1 & 51.8 & 69.9 & 80.3\\
\footnotesize PointNet & 83.7 & 83.4 & 78.7 & 82.5 & 74.9 & 89.6 & 73.0 & 91.5 & 85.9 & 80.8 & 95.3 & 65.2 & 93.0 & 81.2 & 57.9 & 72.8 & 80.6\\ 
\footnotesize PointNet++\cite{qi2017pointnet++} &85.1& 82.4& 79.0 &87.7& 77.3& 90.8 &71.8& 91.0& 85.9&83.7& 95.3& 71.6& 94.1 &81.3 &58.7& 76.4 &82.6\\
\footnotesize P2Sequence\cite{2018Point2Sequence} & 85.1 & 82.6 & 81.8 & 87.5 & 77.3 & 90.8 &77.1 & 91.1 & 86.9 & 83.9 & 95.7 & 70.8 & 94.6 & 79.3 & 58.1 & 75.2 & 82.8\\
\footnotesize DGCNN & 85.2 & 84.0 & 83.4 & 86.7 & 77.8 & 90.6 & 74.7 & 91.2 & 87.5 & 82.8 & 95.7 & 66.3 & 94.9 & 81.1 & 63.5 & 74.5 & 82.6\\
\hline
\hline
\multicolumn{18}{c}{Unsupervised Transfer Learning} \\
\footnotesize LGAN \cite{LGAN} &57.0& 54.1&48.7 &62.6& 43.2& 68.4 &58.3& 74.3& 68.4&53.4& 82.6&18.6& 75.1 &54.7 &37.2& 46.7 &66.4\\
\footnotesize MAP-VAE \cite{MAP-VAE} & 68.0 & 62.7 & 67.1 & 73.0& 58.5 & 77.1 &67.3 & 84.8 & 77.1 & 60.9 & 90.8 & 35.8 & 87.7& 64.2& 45.0 &60.4 & 74.8\\
\footnotesize Graph-TER\cite{gao2020graphter} & 81.9&81.7 & 68.1 & 83.7& 74.6 & 88.1 &68.9 & 90.6 &86.6 & 80.0 & 95.6 & 56.3 & 90.0& 80.8 & 55.2 & 70.7& 79.1\\
\footnotesize MID-FC \cite{wang2020unsupervised} & 84.2 & 80.4 & \textbf{82.5} & \textbf{89.0} & \textbf{80.0} & 89.9& \textbf{80.7} &90.5 &85.7 & 77.8 & \textbf{95.9} & \textbf{73.4} & 94.8& 81.1 & 56.7 & \textbf{81.8} &82.4\\
\footnotesize Ours-DGCNN & \textbf{85.0} & \textbf{83.5} & 82.4 & 86.9 & 77.9 & \textbf{90.4} & 75.6 & \textbf{91.0} & \textbf{86.9} & \textbf{81.0} & 95.1 & 68.9 & \textbf{94.7} & \textbf{81.4}& \textbf{62.5} & 73.1 & \textbf{82.7}\\
\hline
\hline
\multicolumn{18}{c}{Supervised fine-tuning} \\
\footnotesize SSC-RSCNN & 85.2 & $-$ & $-$ & $-$ &$-$ &  $-$& $-$ & $-$ & $-$ & $-$ & $-$ & $-$ & $-$ &$-$  & $-$ & $-$ &$-$ \\
\footnotesize Self-Sup \cite{sauder2019self} & 85.3&84.1 &84.0 & 85.8 & 77.0& 90.9 & 80.0 &91.5 &87.0  & 83.2 & 95.8 &71.6  & 94.0 &  82.6&60.0  &77.9 & 81.8 \\
\footnotesize PointDis \cite{PointDist} & 85.3 & $-$ & $-$ & $-$ &$-$ &  $-$& $-$ & $-$ & $-$ & $-$ & $-$ & $-$ & $-$ &$-$  & $-$ & $-$ &$-$ \\
\footnotesize OcCo \cite{wang2021unsupervised} & 85.5 & 84.4 & 77.5& 83.4 &77.9& 91.0&75.2 & 91.6 & 88.2 & \textbf{83.5} & 96.1 & 65.5 & 94.4 &79.6  &58.0 & 76.2 &82.8 \\
\footnotesize MID-FC \cite{wang2020unsupervised} & 85.5 & 83.6& \textbf{82.9} & \textbf{91.3}&81.6 & 90.4& \textbf{81.5} & \textbf{91.8} & 87.1 & 79.3 & 95.7& 68.7& \textbf{95.2} &83.6 & \textbf{68.3} & \textbf{82.7} &83.2 \\
\footnotesize Ours-DGCNN & \textbf{85.6} & \textbf{84.7} & 80.9 & 86.2 &\textbf{82.3} &  \textbf{91.3}& 76.2 & 91.2 & \textbf{89.6} & 82.1 & \textbf{96.7} & \textbf{68.9} & 93.7 &\textbf{83.8} & 65.3 & 79.5 &\textbf{83.1}\\
\hline
\end{tabular}  
\end{center}  
\caption{We present the part segmentation evaluation results on ShapeNet Part, where mIoU refers to mean Intersection-over-Union measure.}  
\label{part_seg}
\end{table*} 

\textbf{Dataset.} We utilize ModelNet40 \cite{20153D} and ModelNet10 \cite{20153D} for shape classification task. We follow the same data split protocols of PointNet-based methods \cite{2018Dynamic} for these two datasets. For ModelNet40, the train set has 9, 840 models and the test set has 2, 468 models, and the dataset consists of 40 categories. For ModelNet10, 3, 991 models are for fine-tuning and 908 models for testing. It contains 10 categories. We follow the experimental configuration \cite{qi2017pointnet}: (1) we uniformly sample 1,024 points from the mesh faces for each model; (2) the point cloud is re-scaled to fit the unit sphere; and (3) the (x,y,z) coordinates and the normal of the sampled points are used in the experiment. During the training process, randomly scaling and perturbing the objects are adopted as the data augmentation strategy in our experiment.

\textbf{Implementation Details.} The global max pooling and average pooling layer are deployed in our classification head to acquire a 1,296-dimensional global feature vector. Three layers of linear projection with dropout ratio of 50\% are used to get the final classification score. 

For unsupervised transfer learning, we fix the parameters of encoder and only train the classification head. During training,
a random translation in [-0.2, 0.2], a random anisotropic scaling in [0.67, 1.5] and a random input dropout were applied to augment the input data. 
During testing, no data augmentation or voting methods were used. The batch sizes were 32, 200 training epochs were used and the initial learning rates were 10$^{-3}$, with a cosine annealing schedule to adjust the learning rate at every epoch.

For supervised fine-tuning, we fully fine-tuned the pre-trained model on ModelNet40 and ModelNet10 datasets. 

\textbf{Results.} The classification results are presented in Table \ref{tab:cls}. When utilizing DGCNN as the encoder, our method outperforms most of previous unsupervised counterparts and the results on ModelNet10 and ModelNet40 are comparable to certain fully-supervised models. Since the pre-training of the encoder is based on different
datasets, the results demonstrate that our framework has a strong generalizability, which is regarded as a significant application of self-supervised representation learning. Notably, most of the classes are unseen by the model during ShapeNet pre-training. Thus, the superior performance further demonstrates that our model has a good ability to generalize to novel classes.

To the best of our knowledge, the most important application of self-supervised learning methods is to make full use of a large number of unlabeled data and boost the performance of supervised learning methods. Thus, we pre-train the encoder with our framework on ShapeNet and fine-tune the weights on downstream classification tasks and compare the results with other supervised fine-tuning methods. As demonstrated in Table \ref{tab:cls}, the proposed method outperforms all other supervised fine-tuning methods on ModelNet10 and ModelNet40 datasets, which justifies the superiority of our UAE.

\begin{figure*}[ht]
    \centering
    \includegraphics[width=\linewidth]{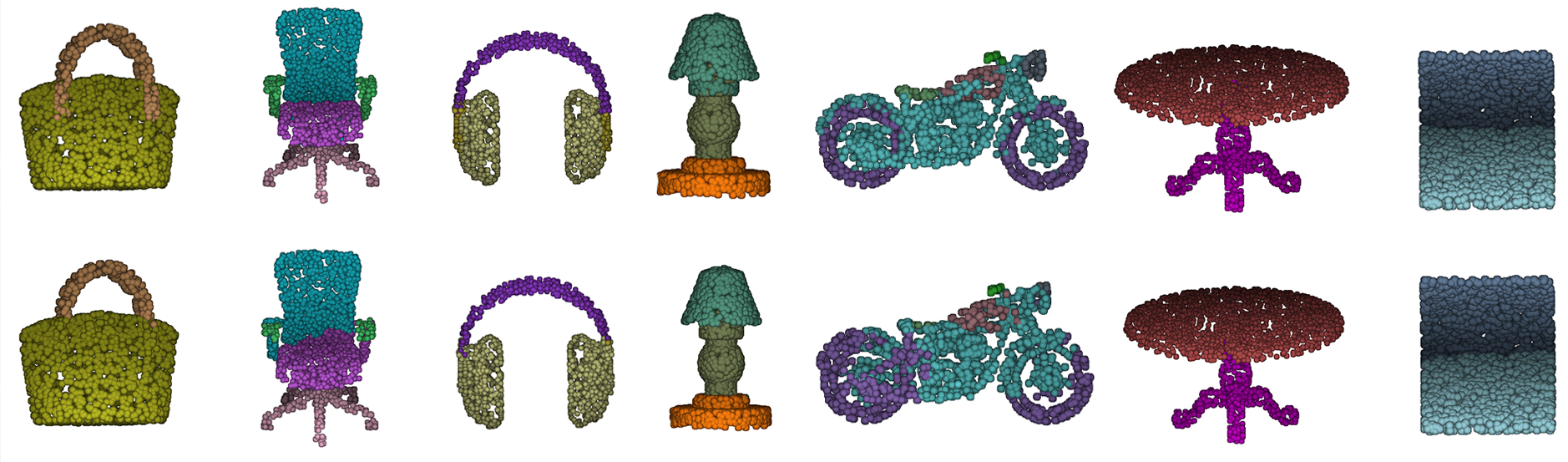}
    \caption{We compare our part segmentation results with the ground truth. From top to bottom: ground truth results and the prediction results of our model.}
    \label{fig:partseg_UAE}
\end{figure*}

\subsection{Part Segmentation}

\textbf{Dataset.} We use the large-scale 3D dataset ShapeNet Parts \cite{shapenetpart} as the experiment bed. ShapeNet Parts contains 16,880 models (14,006 models are used for training, and 2874 models are used for testing), each of which is labeled with two to six parts, and the entire dataset has 50 different part labels. We sample 2,048 points from each model as input, with a few pointsets having six labeled parts. We directly adopt the same train–test split strategy similar to DGCNN \cite{2018Dynamic} in our experiment.

\textbf{Implementation Details.} The global max pooling and average pooling layer are also deployed to acquire a 1,296-dimensional global feature vector. And we design a segmentation head to propagate these features to each point hierarchically. Like shape classification task, three layers of linear projection with dropout ratio of 50\% are also used to get the final classification score of each point. 

We optimize our networks via SGD with batch size 32. The learning rate of unsupervised transfer learning setting starts from 10$^{-2}$ and decreases to 10$^{-4}$, and the learning rate of supervised fine-tuning setting decays from 10$^{-1}$ to 10$^{-3}$.

For unsupervised transfer learning, we fix the parameters of encoder and only train the segmentation head. For supervised fine-tuning, we fully fine-tuned the pre-trained model on ShapeNet Part dataset.

\textbf{Results.}
As shown in Table \ref{part_seg}, the proposed UAE outperforms all other unsupervised point cloud learning models by a large
margin, which indicates that our pre-trained model captures more effective semantic information that can transfer well to downstream segmentation tasks. Some results are visualized in Figure \ref{fig:partseg_UAE}. We also conduct the object part segmentation experiments under supervised fine-tuning strategy and make comparisons with previous excellent models, such as PointDis, MID-FC and OcCo in Table \ref{part_seg}. As shown, our UAE model (Ours-DGCNN) fine-tuned on 100\% labeled samples achieves state-of-the-art performance. Compared to the supervised learning framework DGCNN, our pre-trained model (Ours-DGCNN) achieves remarkable performance improvements, which demonstrates the advantage of our pre-training strategy. We can conclude that pre-training with our
framework on unlabeled data significantly boosts the performance and can be regarded as a strong initializer for supervised models, which is a critical application of self-supervised learning.

\subsection{Point Cloud Upsampling}
\begin{figure*}
    \centering
    \includegraphics[width=\linewidth]{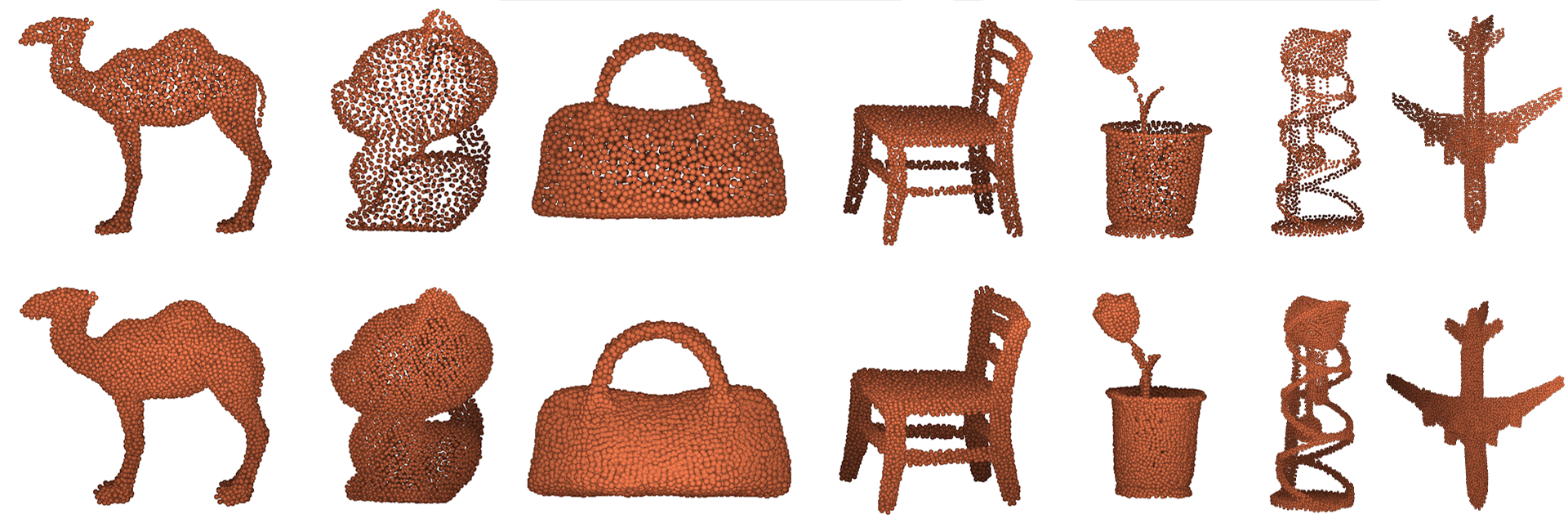}
    \caption{We present the point set upsampling (4x) results. From top to bottom, the input and output are illustrated.}
    \label{fig:upsample}
\end{figure*}
Raw point clouds data acquired from depth cameras or LiDAR scanners are often sparse, noisy and non-uniform, which hinders shape analysis and 3D reconstruction. Point cloud upsampling is thus significant to the
subsequent CAD applications such as analysis, rendering or general processing. We evaluate our UAE on the point cloud upsampling task over the PU-GAN's dataset.

\textbf{Dataset.} For point cloud upsampling task, we use PU-GAN's dataset \cite{PUGAN} as the experiment bed. The PU-GAN's dataset has 147 CAD models that were collected from the released datasets of PU-Net \cite{PUNet} and MPU \cite{MPU}, as well as from the Visionair repository \cite{visi}, covering a rich variety of objects, ranging from simple and smooth models (e.g., Icosahedron) to complex and high-detailed objects (e.g., Statue). Following previous work \cite{PUGAN}, we randomly select 120 models for training and use the rest for testing.

\textbf{Implementation Details.}  We demonstrate a pre-training
strategy to evaluate whether the unsupervised pre-training with our
model helps improve the performance. We first pre-train our UAE on ShapeNet dataset in an unsupervised fashion, and then employ
the learned representation from the pre-trained encoder as an initialization. Eventually, we build our upsampling version of UAE where we only replace the feature extractor of PU-GAN with our pre-trained encoder. We evaluate the effectiveness of our model by comparing the results of PU-GAN with our initialization and other upsampling models in a supervised fashion.

For training, we randomly sample 512 points from each point cloud in the PU-GAN's dataset and upsample them to 2,048 points. For testing, we randomly sample 2048 points and upsample them to 8196 points. The quality of the upsampled point cloud is measured by the CD, HD and P2F between the original and upsampled point clouds.

\textbf{Results.} We qualitatively and quantitatively compared the unsupervised pre-training with our method with the randomly initialized PU-GAN and other state-of-the-art point set upsampling methods: EAR \cite{EAR}, PU-Net \cite{PUNet}, MPU \cite{MPU} and PU-GCN \cite{pugcn}. Table \ref{tab:upsample} shows the quantitative comparison results. Our method (Ours (transfer)) achieves the lowest value consistently for most evaluation metrics in an unsupervised manner. Particularly, the supervised fine-tuning with our method (Ours (fine-tune)) outperforms all previous point set upsampling models. The gain is significant in 4$\times$ upsamling ratio which means that our pre-trained model can capture more effective semantic information. Figure \ref{fig:upsample}
shows the point set upsampling results on PU-GAN's dataset.

\begin{table}[tb]
    \centering
    \begin{tabular}{lcccc}
     \hline
\multirow{2}*{Model} & \multirow{2}*{Supervised} & P2F & CD &HD\\
 &  & (10$^{-3}$) &(10$^{-3}$)&(10$^{-3}$)\\
\hline
\hline
EAR \cite{foldingnet} & yes &5.82 & 0.52& 7.37\\
PU-Net \cite{PUNet} & yes& 6.84 & 0.72&8.94 \\
MPU \cite{MPU} & yes& 3.96 & 0.49 & 6.11 \\
PU-GAN \cite{PUGAN}  & yes& 2.33 & 0.28 &4.61 \\
PU-GCN \cite{pugcn}  & yes& 2.72 & 0.25 &\textbf{1.88} \\
\hline
Ours (transfer)  & no& 2.25 & 0.24 &4.35 \\
Ours (finetune)  & yes& \textbf{2.16} & \textbf{0.22} &4.28 \\
\hline
    \end{tabular}
    \caption{We present quantitative comparisons of our method with the upsampling SOTA approaches.}
    \label{tab:upsample}
\end{table}

\subsection{Ablation Study}
 \begin{table}[htbp]
\centering
\begin{center}  
\begin{tabular}{ccc}
\hline
Sampling strategy  & ModelNet40 & ShapeNet  \\
\hline
Farthest point sampling & 92.64        & 84.83    \\
Local sampling  & 92.46        & 85.04    \\
Random sampling  & \textbf{93.27}& \textbf{85.62}    \\
\hline
\end{tabular}
\end{center}  
\caption{Supervised finetuning results (\% and mIoU) on ModelNet40 and ShapeNet Part with different sampling strategies (the subsampling ratio is 12.5\%), where we can see that the random sampling strategy works the best.}
\label{ablation:strategy}
\end{table}
\begin{table}[tb]
\centering
\begin{center}  
\begin{tabular}{ccc}
\hline
Subsampling ratio ($r$)  & ModelNet40 & ShapeNet  \\
\hline
5.0\%  & 91.58           & 83.94                \\
12.5\% & \textbf{93.27}& \textbf{85.62}\       \\
25.0\%   & 92.69          & 85.22\\
50.0\%   & 92.56          & 84.94\\
100.0\%   & 92.19          & 84.76\\
\hline
\end{tabular}
\end{center}  
\caption{Supervised finetuning results (\% and mIoU) on ModelNet40 and ShapeNet Part datasets with different subsampling rates. }
\label{ablation:ratio}
\end{table}
\begin{figure}[t]
    \centering
    \includegraphics[width=\linewidth]{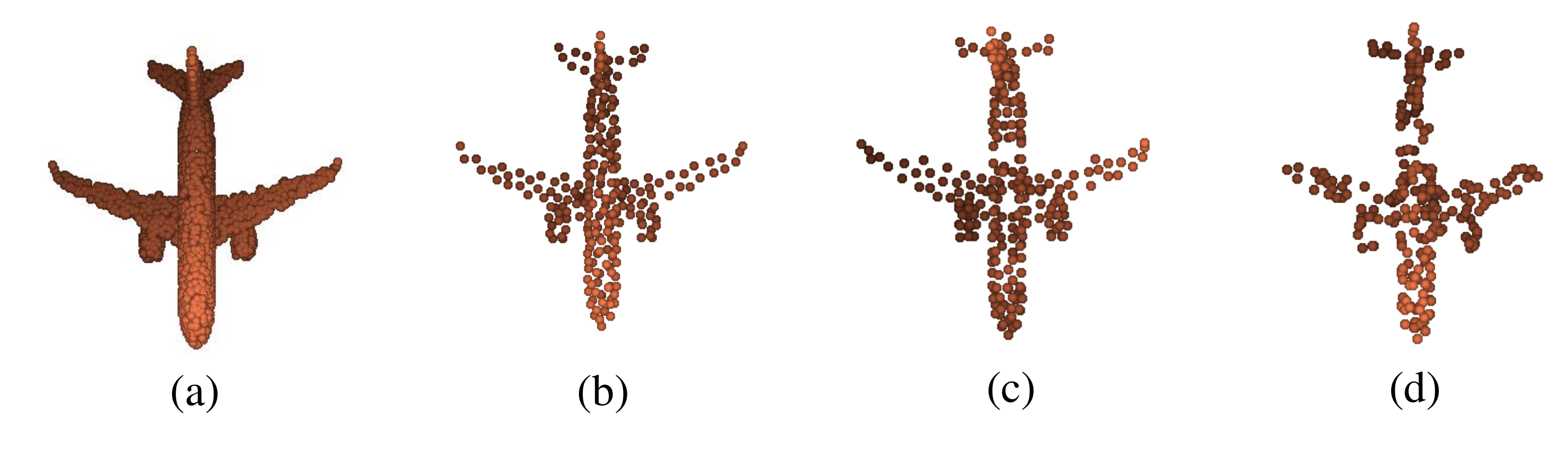}
    \caption{Various subsampling strategies. We subsample 256 points from original 2048 points (a) by perfoming FPS (b), LS (c) and random sampling (d). Form this figure, we can see that random sampling retains less geometric details.}
\label{subsample}
\end{figure}
\begin{table}[t]
\centering
\begin{center}  
\begin{tabular}{ccc}
\hline
Loss function  & ModelNet40 & ShapeNet  \\
\hline
CD & 92.13& 84.53\\
EMD & 92.26& 84.71\\
EMD + RL & 93.18& 85.54\\
CD+RL (Ours) & \textbf{93.27}& \textbf{85.62}\\
\hline
\end{tabular}
\end{center}  
\caption{Ablation study on various loss functions. CD: replace our joint loss function with chamfer distance loss in our architecture. EMD: replace the joint loss function with Earth's moving distance loss in our architecture. EMD + RL: replace our joint loss function with Earth's moving distance and the repulsion losses.}
\label{ablation:loss}
\end{table}

The following studies, which are used to investigate the determining factors of our framework, are conducted on both ModelNet40 and ShapeNetPart datasets.

\textbf{Impact of subsampling strategy.}
To examine the effectiveness of various subsampling strategies, we conduct a detailed ablation on shape classification and part segmentation tasks. Table \ref{ablation:strategy} presents the results with three types of subsampling methods. We observe that the random sampling achieves the best performance, improving by 0.63\%/0.79\% on average over Farthest Point Sampling (FPS), and 0.81\%/0.58\% over local sampling (LS). The reason is that FPS and LS retain too many geometric details (see Figure \ref{subsample}), resulting in the created task being easily solved by extrapolation from neighboring points. In contrast, the task created by random sampling with a low sampling ratio is harder than that of FPS and LS, which provides a more challenging goal for our model.
Moreover, random sampling uniformly selects $rN$ points from the original $N$ points. Its computational complexity is $\mathcal{O}(1)$, which is agnostic to the total number of input points.
Compared with FPS and LS, random sampling has the highest computational efficiency, regardless of the scale of input point clouds \cite{randla}. 
Based on these characteristics in both computational time and memory, we can conclude that our UAE is suitable for training very large models.

\textbf{Effect of subsampling ratio.}
We conduct an ablation study to analyze the setting of subsampling ratio $r$. The results are shown in Table \ref{ablation:ratio}. The best performance is achieved when $r$ is set to 12.5\%. When the ratio becomes smaller, the lack of key points will result in a performance decline. On the contrary, when the ratio increases, a lot of noises at the local boundary lead to the deformation of feature extraction ability, and then reduces the accuracy of the model. Meanwhile, the task with high subsampling ratio (25\% or 50\%) can be easily solved by our UAE, which is not conducive to capture high-level semantic information.

\textbf{Impact of loss function.}
We also investigate the options of loss functions. The results are shown in Table \ref{ablation:loss}. From this table, we can see that compared with the CD loss, EMD loss achieves better results (+0.13\%/+0.18\%) because it can better encourage the output points to
be located close to the underlying object surfaces \cite{PUNet}. However, in Table \ref{ablation:loss}, the results of the two tasks are close, the time complexity of EMD loss is $\mathcal{O}(N^2)$ \cite{PUGAN}, which means that it needs more pretraining time, especially when $N$ is very large. Furthermore, we can see that the performance of the model is significantly improved after adding repulsion loss, especially under the combination of “CD + RL” where the results from the output points are marked semantically and distinguished in spatial position to reduce noise.

\section{Conclusion and Future work}
In this work, we presented UAE, a new framework for self-supervised point cloud learning. UAE learns high-level semantic information by upsampling a sparse point cloud uniformly within a simple and effective framework.
We showed state-of-the-art results of our method on various downstream tasks including shape classification, object segmentation and point cloud upsampling. In the future, it would be interesting to further exploit other applications, 
and take one closer step to the harsh real-world setting, i.e., limited annotations. In addition,
we will continue to study the point cloud pre-training methods on large-scale datasets, and focus on finding an efficient way to take advantage of the large-scale point data.

\section{Acknowledgements}
This work was partially supported by the Zhejiang Provincial Natural Science Foundation of China (LGF21F20012), the National Natural Science Foundation
of China (No.61602139), and the Graduate Scientific Research Foundation of Hangzhou Dianzi University (CXJJ2021082, CXJJ2021083).

{\small
\bibliographystyle{ieee}
\bibliography{egbib}
}

\end{document}